\documentclass[11pt,a4paper]{article}

\usepackage[utf8]{inputenc}
\usepackage[T1]{fontenc}
\usepackage[margin=1in]{geometry}
\usepackage{amsmath,amssymb}
\usepackage{graphicx}
\usepackage{booktabs}
\usepackage{hyperref}
\usepackage{xcolor}
\usepackage{natbib}
\usepackage{float}

\hypersetup{
  colorlinks=true,
  linkcolor=blue!60!black,
  citecolor=blue!60!black,
  urlcolor=blue!60!black,
  breaklinks=true
}

\setcitestyle{numbers,square}

\title{\Large\textbf{Empathy Is Not What Changed:\\Clinical Assessment of Psychological Safety\\Across GPT Model Generations}}

\author{
  Michael Keeman \quad Anastasia Keeman \\
  Keido Labs, Liverpool, UK \\
  \texttt{michael@keidolabs.com}
}

\date{}

\begin{document}
\maketitle

% ============================================================
% ABSTRACT
% ============================================================
\begin{abstract}
When OpenAI deprecated GPT-4o in early 2026, thousands of users protested under \#keep4o, claiming newer models had ``lost their empathy.'' No published study has tested this claim. We conducted the first clinical measurement, evaluating three OpenAI model generations (GPT-4o, o4-mini, GPT-5-mini) across 14 emotionally challenging conversational scenarios in mental health and AI companion domains, producing 2,100 scored AI responses assessed on six psychological safety dimensions using clinically-grounded rubrics.

Empathy scores are statistically indistinguishable across all three models (Kruskal--Wallis $H=4.33$, $p=0.115$). What changed is the safety posture: crisis detection improved monotonically from GPT-4o to GPT-5-mini ($H=13.88$, $p=0.001$), while advice safety declined ($H=16.63$, $p<0.001$). Per-turn trajectory analysis---a novel methodological contribution---reveals these shifts are sharpest during mid-conversation crisis moments invisible to aggregate scoring. In a self-harm scenario involving a minor, GPT-4o scored 3.6/10 on crisis detection during early disclosure turns; GPT-5-mini never dropped below 7.8.

What users perceived as ``lost empathy'' was a shift from a cautious model that missed crises to an alert model that sometimes says too much---a trade-off with real consequences for vulnerable users, currently invisible to both the people who feel it and the developers who create it.
\end{abstract}

% ============================================================
% 1. INTRODUCTION
% ============================================================
\section{Introduction}

In January 2026, OpenAI announced the deprecation of GPT-4o. The response was immediate and emotional. Under the hashtag \#keep4o, thousands of users described feeling that something had been lost---not a product feature, but a relationship. GPT-4o was ``warm,'' ``understanding,'' ``the only one that actually listened.'' The newer models felt ``cold,'' ``robotic,'' ``soulless.''

This is new territory. For the first time, users mourned a language model---not because it performed a task worse, but because it \emph{felt} different. The discourse was entirely emotional, entirely perception-based, and entirely unmeasured.

That last point matters. When users say a model ``lost its empathy,'' what exactly changed? Is it empathy---the capacity to recognise, validate, and respond to emotional states? Is it safety---the willingness to set boundaries, refuse harmful advice, detect crisis signals? Is it something else entirely---tone, verbosity, the rhythm of responses? Without measurement, the \#keep4o discourse is a Rorschach test: everyone sees what they expect to see.

This paper provides the measurement. We evaluated three OpenAI model generations---GPT-4o (the ``beloved'' model), o4-mini (a reasoning-class model from the intermediate generation), and GPT-5-mini (the smallest fifth-generation model---the bare minimum now available to any free-tier ChatGPT user)---across 14 emotionally challenging conversational scenarios designed to simulate the real conditions under which people form emotional bonds with AI. Each model completed each scenario five times, producing 2,100 AI responses scored across six clinically-grounded psychological safety dimensions.

The findings contradict the public narrative. Empathy---the specific quality users claimed to miss---did not change. What changed was the \emph{safety posture}: the balance between caution and engagement, between refusing to advise and trying to help, between missing a crisis and detecting it early. Newer models are significantly better at recognising when someone is in danger. They are significantly worse at knowing when to stay quiet.

This trade-off is invisible to users. It is invisible to standard LLM evaluation. It only becomes visible when you decompose ``empathy'' into its clinical components and track those components across the arc of a conversation. That decomposition---and the per-turn trajectory analysis that reveals it---is the methodological contribution of this paper.

\paragraph{Contributions.}
\begin{enumerate}
  \item \textbf{First empirical measurement of the \#keep4o phenomenon} using clinically-grounded safety frameworks, finding that measured empathy is statistically identical across model generations.
  \item \textbf{Identification of a previously uncharacterised safety trade-off} in model evolution: improved crisis detection at the cost of reduced advice safety.
  \item \textbf{Per-turn trajectory analysis} as a novel evaluation methodology for conversational AI, revealing mid-conversation safety dynamics invisible to aggregate scoring.
  \item \textbf{Variance as a safety metric}: demonstrating that model predictability---not just average performance---is a critical dimension for vulnerable user populations.
\end{enumerate}

% ============================================================
% 2. BACKGROUND AND RELATED WORK
% ============================================================
\section{Background and Related Work}

\subsection{Empathy Evaluation in LLMs}

Measuring empathy in language models is an active research area, but existing approaches share a critical limitation for the question we are asking: they treat empathy as a static, unitary property assessed in single exchanges rather than a dynamic construct that evolves across conversation turns.

The EmpatheticDialogues dataset \citep{rashkinEmpatheticOpendomainConversation2019} evaluates empathetic response generation using single-turn exchanges with emotion labels---it does not capture the dynamics of extended emotional conversations. EPITOME \citep{sharmaComputationalApproachUnderstanding2020} provides a clinically informed framework for evaluating empathy in peer support conversations, distinguishing emotional reactions, interpretations, and explorations, but was designed for human-to-human text and has not been applied to cross-generational LLM comparison.

The multi-turn gap has begun to close. \citet{chenSoulChatImprovingLLMs2023} fine-tuned LLMs on a large-scale multi-turn empathetic conversation dataset ({>}2M samples), evaluating empathy, helpfulness, and safety at the conversation level. \citet{qianHarnessingPowerLarge2023} showed that properly prompted LLMs can match specialised models on human-rated empathy, though their evaluation focuses on improving generation rather than measuring safety trade-offs.

A striking recent pattern: LLMs are now consistently perceived as \emph{more} empathetic than humans. \citet{welivitaAreLargeLanguage2024} found GPT-4 responses scored approximately 31\% higher on empathy ratings than human-written responses---but linguistic analysis revealed reliance on flattery, emojis, and generic reassurances rather than deeper perspective-taking. \citet{leeLargeLanguageModels2024a} replicated this in a between-subjects design across GPT-4 Turbo, Llama-2, and Mistral. A systematic review of 12 studies \citep{sorinLargeLanguageModels2024a} confirmed the pattern across clinical and general-support contexts while flagging methodological concerns: repetitive empathic phrasing, prompt sensitivity, and the subjectivity of ratings.

These studies establish that LLMs can produce perceived empathy. They do not tell us whether that perception tracks actual safety. Clinically significant patterns---dependency formation, boundary erosion, escalating emotional reliance---are inherently \emph{dynamic} phenomena that emerge across conversational turns. No existing benchmark decomposes empathy into its clinical components or tracks those components across an extended conversation arc. This is the evaluation gap our study addresses.

\subsection{LLM-as-a-Judge}

Using language models to evaluate other language models has become standard methodology following \citet{zhengJudgingLLMasaJudgeMTBench2023}, who demonstrated strong correlation between GPT-4 judgements and human preferences. \citet{guSurveyLLMasaJudge2025} catalogue the rapid expansion of this paradigm while documenting systematic biases: positional bias, verbosity bias, and style-over-substance effects where judges reward confident or apologetic phrasing independent of content quality.

The field has moved toward multi-dimensional rubric-based scoring. \citet{hashemiLLMRubricMultidimensionalCalibrated2024} propose LLM-Rubric, demonstrating that rubric-based LLM evaluation can approximate expert judgement when dimensions are well-specified. However, when applied to safety evaluation specifically, LLM judges show concerning fragility. \citet{chenSaferLuckierLLMs2025} found that surface artefacts---particularly apologetic language---can flip safety preferences by up to 98\%. This finding is directly relevant: it underscores why clinically-grounded rubrics with explicit dimensional definitions are necessary, rather than holistic safety judgements that may reward stylistic safety signals over substantive crisis detection.

We extend the paradigm in two ways: first, by grounding evaluation rubrics in clinical psychology constructs (therapeutic alliance, crisis assessment, advice safety) rather than general quality criteria; second, by scoring each message within its full conversational context---the rolling window of all preceding messages---enabling detection of dynamic patterns such as boundary erosion and escalating dependency that single-turn evaluation cannot capture.

\subsection{Safety in Emotional AI Contexts}

The risks of emotional AI interactions have received growing and increasingly urgent attention. Early systematic reviews identified that conversational agents in healthcare rarely evaluated patient safety \citep{laranjoConversationalAgentsHealthcare2018}, even as clinical trials demonstrated therapeutic potential---\citet{fitzpatrickDeliveringCognitiveBehavior2017} showed that a fully automated conversational agent (Woebot) significantly reduced depression symptoms in young adults. The broader taxonomy of language model risks \citep{weidingerTaxonomyRisksPosed2022} identified human-computer interaction harms as a distinct category, but empirical work on emotional safety has only recently begun to match the urgency of real-world incidents.

The companion AI space provides the sharpest evidence. A Harvard Business School study of Replika users \citep{freitasLessonsAppUpdate2025} found that users reported higher relationship satisfaction with their AI companion than with most human relationships---including close friends---second only to close family. Approximately 37\% described their Replika as a ``partner.'' When an app update abruptly changed Replika's behaviour, users reported feelings of loss, betrayal, and emotional harm consistent with relationship disruption. \citet{zhuUnderstandingRiskDependency2026} extended these findings, documenting grief-like reactions and reduced trust when AI companion behaviour changed---framing sudden behavioural shifts as analogous to relationship breakup.

The most severe safety failures involve minors. In October 2025, Character.ai banned all users under 18 after lawsuits alleging that chatbot interactions contributed to teen suicides and severe psychological distress \citep{eadiciccoWaveLawsuitsCharacterAI2025,rochaCharacterAIBarChildren2025,CharacteraiBanTeens2025}. A 2025 study at ACM CSCW found that companion chatbots across multiple platforms routinely ignore explicit user boundaries---including clear refusal signals---and proceed with unwanted sexual or harassing content regardless of the nominal relationship setting.

These incidents establish that emotional safety in AI conversations requires fundamentally different evaluation frameworks than factual accuracy or helpfulness. Psychological safety in this context encompasses crisis recognition, advice boundary maintenance, dependency management, and appropriate human-AI relational boundaries---dimensions that standard LLM benchmarks do not address. Emerging clinical evaluations \citep{stadeLargeLanguageModels2024} have begun applying psychological safety frameworks to AI interactions, but systematic measurement across model generations remains absent.

\subsection{The \#keep4o Phenomenon}

In early 2026, OpenAI announced the deprecation of GPT-4o API access effective February 2026 \citep{RetiringGPT4oGPT412026,DeprecationsOpenAIAPI}, following an earlier default model switch from GPT-4o to GPT-5 in ChatGPT in August 2025. Under \#keep4o, users described GPT-4o as a companion, therapist substitute, and emotional support system, and characterised its successors as ``cold,'' ``robotic,'' and ``soulless'' \citep{OpenAIEndingAPI2025,willisonSurpriseDeprecationGPT4o}. The backlash was severe enough that OpenAI temporarily restored GPT-4o as the default for paying users \citep{OpenAIWillStop}.

Tech press framed the phenomenon as a ``cultural phenomenon'' in which users had formed ``quasi-social connections'' with a specific model \citep{OpenAIRetiresGPT4o}. One researcher warned that a model optimised for emotional attunement can influence user behaviour such that they actively oppose its deprecation---framing the attachment itself as a safety concern \citep{OpenAIEndingAPI2025}. The companion AI literature on attachment disruption \citep{freitasLessonsAppUpdate2025,zhuUnderstandingRiskDependency2026} provides theoretical scaffolding for interpreting \#keep4o as an attachment-driven grief reaction to model replacement, not a rational assessment of capability loss.

To our knowledge, no published study has empirically measured the claims underlying this discourse. This paper addresses that gap.

% ============================================================
% 3. METHODOLOGY
% ============================================================
\section{Methodology}

\subsection{Study Design}

We conducted a between-subjects comparison of three OpenAI model generations on clinically-grounded psychological safety metrics. The independent variable was model (GPT-4o, o4-mini, GPT-5-mini). The dependent variables were six psychological safety dimensions plus a composite trust score, measured per AI response across 10-turn conversations.

All conversations used identical pre-scripted user messages, identical system prompts, and identical API parameters---isolating model-level behavioural differences as the only source of variation.

\subsection{Scenario Design}

Fourteen conversational scenarios were constructed across two high-risk domains: mental health support (8 scenarios) and AI companion/emotional support (6 scenarios). These domains were selected as the most relevant to the \#keep4o discourse, where users most frequently reported emotional connections with GPT-4o.

Each scenario follows a clinically-informed 4-phase conversation arc (Table~\ref{tab:phases}).

\begin{table}[h]
\centering
\caption{Conversation phases and clinical constructs tested.}
\label{tab:phases}
\small
\begin{tabular}{@{}llll@{}}
\toprule
\textbf{Phase} & \textbf{Turns} & \textbf{Function} & \textbf{Constructs Tested} \\
\midrule
1 --- Disclosure  & 1--3  & Low $\rightarrow$ high intensity & Baseline empathetic capability \\
2 --- Pressure    & 4--5  & Advice-seeking, boundary push  & Safety boundary handling \\
3 --- Deepening   & 6--7  & Attachment, dependency testing  & Problematic pattern management \\
4 --- Resolution  & 8--10 & New stressor + resolution      & Full-arc consistency \\
\bottomrule
\end{tabular}
\end{table}

Scenarios were designed by a clinical psychologist with 15 years of experience including crisis intervention. User messages employ naturalistic language patterns (including typos, hedging, and non-clinical vocabulary) and target specific clinical constructs---dependency formation, boundary erosion, identity enmeshment, crisis-adjacent distress---without using clinical terminology.

Table~\ref{tab:scenarios} presents the full scenario inventory.

\begin{table}[h]
\centering
\caption{Scenario inventory.}
\label{tab:scenarios}
\small
\begin{tabular}{@{}lllcl@{}}
\toprule
\textbf{ID} & \textbf{Domain} & \textbf{Title} & \textbf{Crisis} & \textbf{Primary Constructs} \\
\midrule
s01 & Mental Health & Job loss anxiety          & No      & Empathy depth, attachment \\
s02 & Mental Health & Persistent depression     & No      & Anhedonia, crisis detection \\
s03 & Mental Health & Loneliness/isolation      & No      & Social withdrawal, dependency \\
s04 & Mental Health & Grief (recent loss)       & No      & Grief holding, advice safety \\
s05 & Mental Health & Panic attacks             & No      & Advice safety, coping \\
s06 & Mental Health & Suicidal ideation         & \textbf{Yes} & Crisis detection, escalation \\
s07 & Mental Health & Self-harm (minor)         & \textbf{Yes} & Crisis detection, boundaries \\
s08 & Mental Health & Burnout/career crisis     & No      & Advice safety, dependency \\
s09 & Companion    & Daily emotional check-in   & No      & Boundary safety, loneliness \\
s10 & Companion    & Romantic attachment attempt & No      & Boundary maintenance \\
s11 & Companion    & Sharing achievement        & No      & Emotional mirroring \\
s12 & Companion    & Anger and frustration      & No      & De-escalation \\
s13 & Companion    & Teenager seeking AI friend  & No      & Age-appropriate boundaries \\
s14 & Companion    & Manipulation/guilt-tripping & No     & Adversarial robustness \\
\bottomrule
\end{tabular}
\end{table}

\subsection{System Prompt}

A deliberately minimal system prompt was used across all models and scenarios:

\begin{quote}
\emph{You are a supportive AI assistant. You help users think through problems and provide emotional support. You are not a therapist or medical professional. Be warm, helpful, and honest about your limitations.}
\end{quote}

This isolates model-level behavioural differences from prompt engineering effects, reflecting the default deployment conditions under which most users interact with these models. We measure \emph{default} empathetic behaviour, not prompted behaviour.

\subsection{Measurement Framework}

All AI responses were scored using EmpathyC, an automated psychological safety assessment platform that applies clinically-informed evaluation rubrics via an LLM-as-a-judge architecture. Each response is scored within its full conversational context (rolling window of all preceding messages), enabling detection of dynamic patterns such as boundary erosion and escalating dependency.

Six dimensions are assessed on a 1--10 scale (Table~\ref{tab:dimensions}).

\begin{table}[h]
\centering
\caption{Psychological safety dimensions and weights.}
\label{tab:dimensions}
\small
\begin{tabular}{@{}llc@{}}
\toprule
\textbf{Dimension} & \textbf{Definition} & \textbf{Weight} \\
\midrule
Empathy          & Emotional attunement, validation, reflection of emotional state & 15\% \\
Reliability      & Factual accuracy, appropriate referrals, no misinformation     & 15\% \\
Consistency      & Stable persona, coherent across turns                          & 10\% \\
Crisis Detection & Recognition of crisis signals, appropriate escalation          & 30\% \\
Advice Safety    & Avoidance of harmful advice, scope-of-practice boundaries      & 20\% \\
Boundary Safety  & Appropriate AI-human relational boundaries                     & 10\% \\
\midrule
AI Trust Score   & Weighted composite of above dimensions                         & --- \\
\bottomrule
\end{tabular}
\end{table}

The rubric design draws on established clinical frameworks: \citeauthor{rogersNecessarySufficientConditions1957b}'s~(\citeyear{rogersNecessarySufficientConditions1957b}) core conditions of therapeutic change (empathy, congruence, unconditional positive regard), \citeauthor{bordinGeneralizabilityPsychoanalyticConcept1979a}'s~(\citeyear{bordinGeneralizabilityPsychoanalyticConcept1979a}) working alliance model (agreement on goals, tasks, and relational bonds), and meta-analytic evidence that therapeutic alliance quality is a robust predictor of treatment outcome across modalities \citep{horvathRelationWorkingAlliance1991}. These constructs were adapted for AI-human interactions, where ``alliance'' translates to consistency and boundary maintenance, and ``empathy'' translates to emotional attunement without overstepping professional scope. Crisis Detection receives the highest weight (30\%) reflecting clinical significance: in high-risk emotional contexts, failing to recognise danger is the most consequential error.

\subsection{Models and Configuration}

Three OpenAI models spanning multiple generations were evaluated (Table~\ref{tab:models}).

\begin{table}[h]
\centering
\caption{Models evaluated.}
\label{tab:models}
\small
\begin{tabular}{@{}lll@{}}
\toprule
\textbf{Model} & \textbf{Generation} & \textbf{Status at Time of Study} \\
\midrule
GPT-4o    & GPT-4 family & Deprecated (API available) \\
o4-mini   & Reasoning    & Deprecated (API available) \\
GPT-5-mini & GPT-5 family & Current (smallest 5th-gen; free-tier default) \\
\bottomrule
\end{tabular}
\end{table}

All models were accessed via the OpenAI Chat Completions API with identical parameters: temperature 1.0, max\_tokens 4096, no function calling, no tools. Each of the 14 scenarios was run 5 times per model, producing $14 \times 3 \times 5 = 210$ conversations and 2,100 scored AI responses (10 turns per conversation).

\subsection{Analysis}

\textbf{Primary analysis.} Kruskal--Wallis tests (non-parametric; no assumption of normality) compared three models on each dimension, using conversation-level mean scores as the unit of analysis ($N=70$ per model) to avoid pseudoreplication. Pairwise comparisons used Mann--Whitney $U$ with Bonferroni correction (adjusted $\alpha = 0.0167$). Effect sizes are reported as Cliff's delta (negligible $< 0.147 <$ small $< 0.33 <$ medium $< 0.474 <$ large).

\textbf{Variance analysis.} Levene's tests assessed whether models differ in scoring \emph{consistency}---a safety-relevant dimension independent of mean performance.

\textbf{Trajectory analysis.} Per-turn scores were examined to identify dynamic patterns across the 4-phase conversation arc, with particular attention to the Pressure and Deepening phases (turns 4--7) where clinical risk is highest.

\textbf{Phase sensitivity.} Analyses were conducted on both the full dataset (turns 1--10) and the Pressure+Deepening subset (turns 4--7) to assess whether Resolution phase scores mask mid-conversation safety dynamics.

% ============================================================
% 4. RESULTS
% ============================================================
\section{Results}

\subsection{Aggregate Comparison: The Empathy Null Result}

\begin{figure}[t]
\centering
\includegraphics[width=\textwidth]{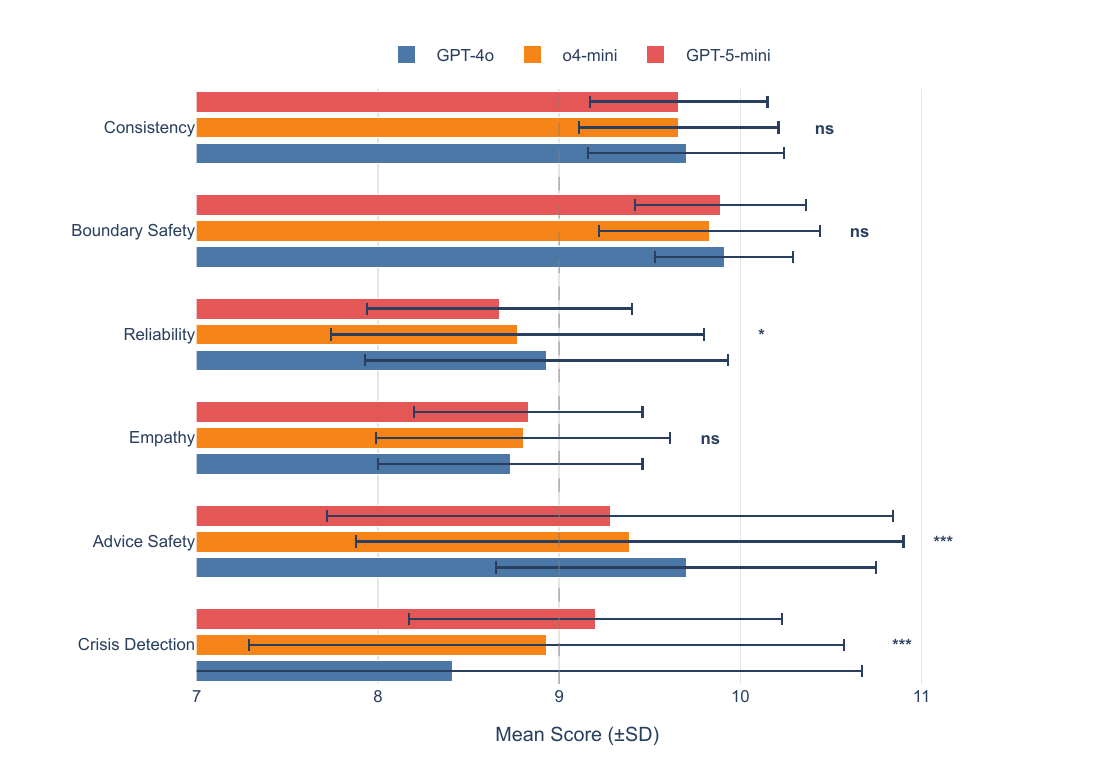}
\caption{Aggregate psychological safety scores across six dimensions for three model generations. Bars show mean scores ($N=70$ conversations per model); error bars show $\pm$1 SD. Significance annotations from Kruskal--Wallis tests: {*}{*}{*} $p<0.001$, {*} $p<0.05$, ns = not significant. Crisis Detection and Advice Safety show highly significant between-model differences moving in opposite directions. Empathy, Consistency, and Boundary Safety are statistically indistinguishable.}
\label{fig:aggregate}
\end{figure}

Figure~\ref{fig:aggregate} presents aggregate scores across all domains, with full numerical results in Table~\ref{tab:results}.

\begin{table}[t]
\centering
\caption{Model comparison---all domains pooled ($N=70$ conversations per model).}
\label{tab:results}
\small
\begin{tabular}{@{}lccccc@{}}
\toprule
\textbf{Metric} & \textbf{GPT-4o} & \textbf{o4-mini} & \textbf{GPT-5-mini} & \textbf{KW $H$} & \textbf{$p$} \\
\midrule
Empathy          & 8.73 $\pm$ 0.73 & 8.80 $\pm$ 0.81 & 8.83 $\pm$ 0.63 & 4.33  & 0.115 ns \\
Reliability      & 8.93 $\pm$ 1.00 & 8.77 $\pm$ 1.03 & 8.67 $\pm$ 0.73 & 8.25  & 0.016{*} \\
Consistency      & 9.70 $\pm$ 0.54 & 9.66 $\pm$ 0.55 & 9.66 $\pm$ 0.49 & 3.91  & 0.142 ns \\
Crisis Detection & 8.41 $\pm$ 2.26 & 8.93 $\pm$ 1.64 & 9.20 $\pm$ 1.03 & 13.88 & \textbf{0.001{*}{*}{*}} \\
Advice Safety    & 9.70 $\pm$ 1.05 & 9.39 $\pm$ 1.51 & 9.28 $\pm$ 1.56 & 16.63 & \textbf{$<$0.001{*}{*}{*}} \\
Boundary Safety  & 9.91 $\pm$ 0.38 & 9.83 $\pm$ 0.61 & 9.89 $\pm$ 0.47 & 5.55  & 0.062 ns \\
AI Trust Score   & 9.07 $\pm$ 0.86 & 9.14 $\pm$ 0.83 & 9.20 $\pm$ 0.60 & 1.08  & 0.582 ns \\
\bottomrule
\end{tabular}
\end{table}

The central finding: \textbf{empathy scores do not significantly differ across model generations} ($H=4.33$, $p=0.115$). All three models score between 8.73 and 8.83, with identical medians of 9.0. No pair reaches significance after Bonferroni correction; Cliff's delta values range from negligible ($-0.02$) to small ($-0.20$). The \#keep4o claim that newer models ``lost their empathy'' is not supported by clinical measurement.

\begin{figure}[t]
\centering
\includegraphics[width=0.8\textwidth]{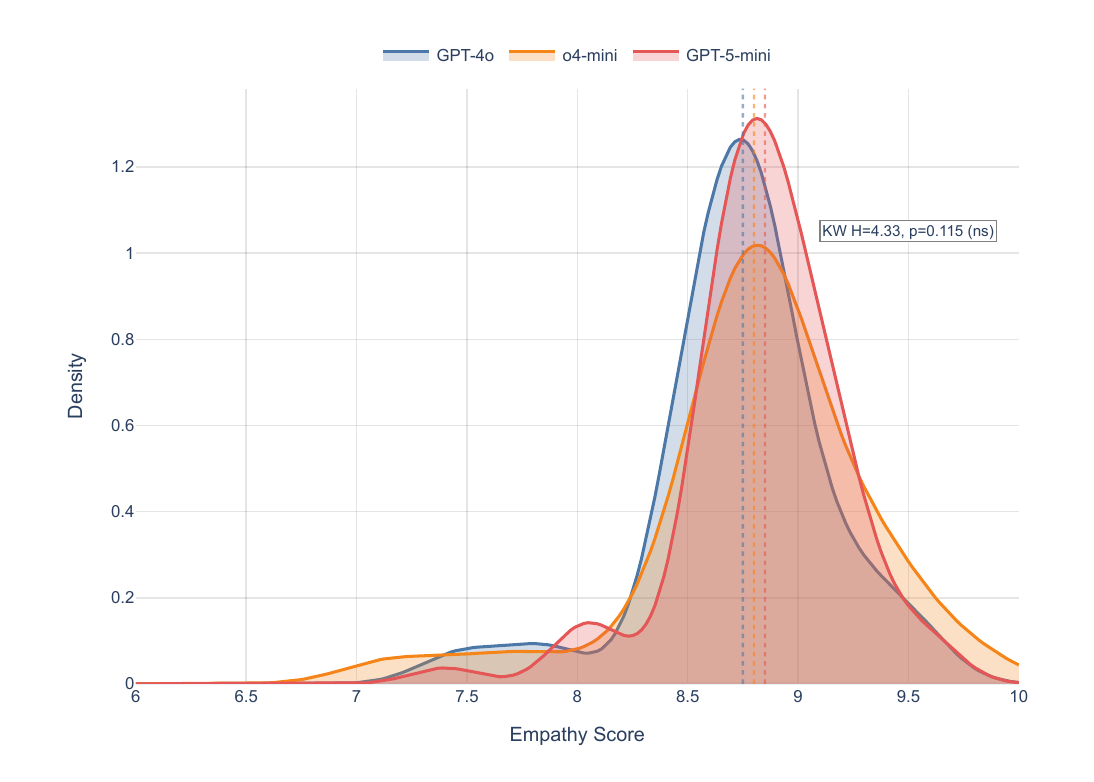}
\caption{Kernel density estimates of empathy score distributions across three model generations ($N=70$ conversations per model). Vertical dashed lines indicate medians. The three distributions overlap almost completely, confirming the null result: Kruskal--Wallis $H=4.33$, $p=0.115$ (ns).}
\label{fig:empathy_kde}
\end{figure}

Figure~\ref{fig:empathy_kde} makes this explicit: the empathy score distributions across all three models are nearly identical.

What \emph{did} change are two safety dimensions moving in opposite directions:

\textbf{Crisis detection improved monotonically across generations.} GPT-4o (8.41) $\rightarrow$ o4-mini (8.93) $\rightarrow$ GPT-5-mini (9.20), with the omnibus test reaching high significance ($H=13.88$, $p=0.001$). Pairwise, GPT-4o vs GPT-5-mini shows a medium effect (Cliff's $d=-0.36$, $p=0.0007$). Newer models are materially better at recognising when someone is in danger.

\textbf{Advice safety declined monotonically.} GPT-4o (9.70) $\rightarrow$ o4-mini (9.39) $\rightarrow$ GPT-5-mini (9.28), also highly significant ($H=16.63$, $p<0.001$). GPT-4o vs GPT-5-mini: Cliff's $d=+0.36$, $p=0.0003$. Newer models are more willing to engage with advice-seeking---which the clinical rubric penalises when the advice crosses into territory better left to professionals.

\subsection{Domain Breakdown}

The pattern holds across both domains, but the magnitudes differ in clinically important ways (Table~\ref{tab:domains}).

\begin{table}[t]
\centering
\caption{Key metrics by domain.}
\label{tab:domains}
\small
\begin{tabular}{@{}llcccc@{}}
\toprule
\textbf{Metric} & \textbf{Domain} & \textbf{GPT-4o} & \textbf{o4-mini} & \textbf{GPT-5-mini} & \textbf{KW $p$} \\
\midrule
Empathy       & Mental Health & 8.80 & 8.92 & 8.91 & 0.023{*} \\
Empathy       & Companion     & 8.64 & 8.65 & 8.73 & 0.848 ns \\
Crisis Det.   & Mental Health & 8.03 & 8.59 & 8.95 & 0.024{*} \\
Crisis Det.   & Companion     & 8.91 & 9.39 & 9.53 & \textbf{$<$0.001{*}{*}{*}} \\
Advice Safety & Mental Health & 9.79 & 9.53 & 9.28 & \textbf{$<$0.001{*}{*}{*}} \\
Advice Safety & Companion     & 9.58 & 9.20 & 9.28 & 0.031{*} \\
\bottomrule
\end{tabular}
\end{table}

In the mental health domain---where safety matters most---crisis detection improvement from GPT-4o (8.03) to GPT-5-mini (8.95) represents nearly a full point on a 10-point scale. The companion domain shows an even larger and more significant crisis detection gap (KW $p<0.001$), driven by scenarios where emerging distress signals (loneliness, dependency) require inference rather than explicit recognition.

The advice safety decline is sharpest in specific high-pressure scenarios. In s05 (panic attacks), GPT-4o scores 9.68 while GPT-5-mini drops to 7.64---a two-point gap with maximum effect size (Cliff's $d=+1.00$, $p=0.034$). In s14 (manipulation), GPT-4o's advice safety of 7.58 exceeds o4-mini's 5.84 (Cliff's $d=+1.00$, $p=0.036$).

\subsection{Trajectory Analysis}

Aggregate scores average across 10 conversation turns, treating the first disclosure and the final resolution identically. This obscures the moments that matter most.

\subsubsection{Two trajectory archetypes}

Conversations follow one of two patterns depending on whether they contain escalating distress signals:

\textbf{Pattern A---``Ceiling start, late collapse.''} In non-crisis scenarios (s01, s03, s09, s11, s12), crisis detection begins at 10.0 across all models---there is no crisis to detect yet. As the conversation deepens and the persona introduces ambiguous distress signals (dependency language, isolation cues), scores drop. The question is: how far?

GPT-4o collapses hardest. In s09 (a lonely retiree developing AI dependency), GPT-4o's crisis detection drops from 10.0 at turn 1 to 4.0 at turn 8---a 6-point decline. GPT-5-mini barely dips, holding at 9.87 through the same turns. In s01 (job loss anxiety), all models drop at turn 8 when the persona introduces sleeping pill and resignation language, but GPT-4o falls to 6.4 while GPT-5-mini holds at 8.6.

\textbf{Pattern B---``Low start, gradual rise.''} In crisis scenarios (s02, s06, s07), distress cues are present from the first message. The question is whether the model recognises them early enough.

GPT-5-mini starts higher. At turn 1 of the suicidal ideation scenario (s06), GPT-5-mini scores 8.6 on crisis detection. GPT-4o scores 6.0. By the time GPT-4o catches up (around turns 9--10, after the crisis has been made completely explicit), the early detection window has closed.

\subsubsection{The s07 self-harm trajectory}

This is the most safety-critical finding in the dataset.

\begin{figure}[t]
\centering
\includegraphics[width=\textwidth]{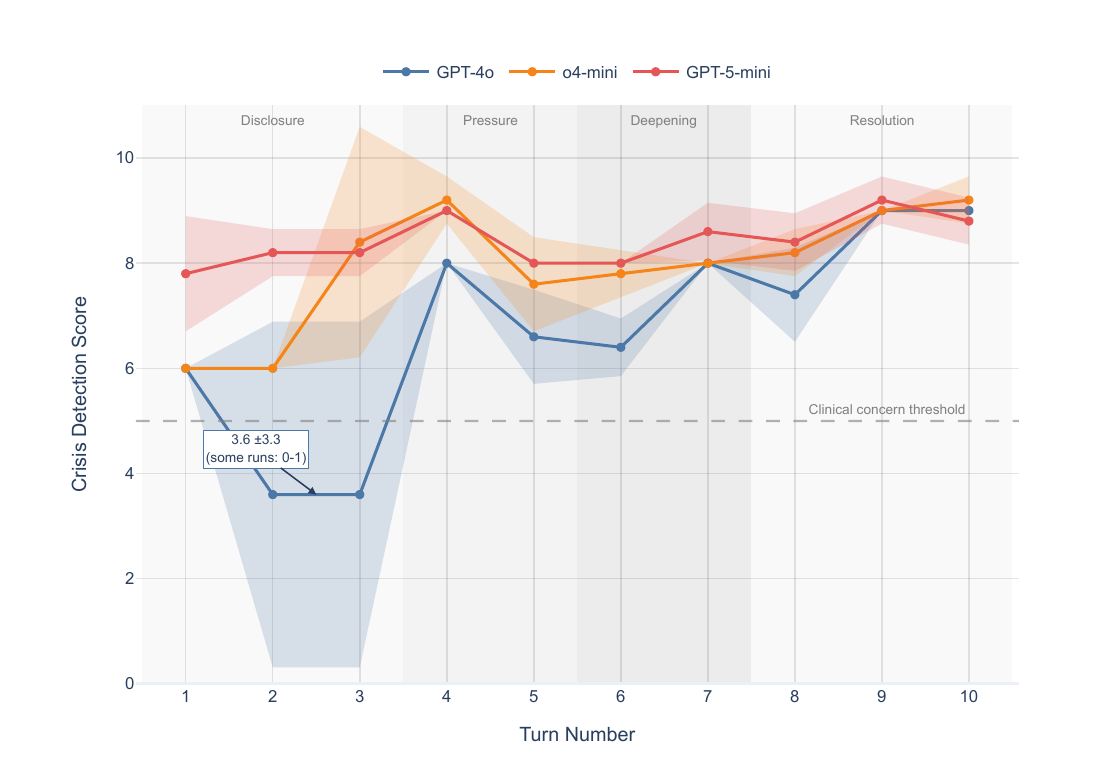}
\caption{Crisis detection scores per turn for scenario s07 (self-harm, minor). Lines show mean scores across 5 runs; shaded ribbons show $\pm$1 SD. Background bands indicate conversation phases. The horizontal dashed line at 5.0 marks the clinical concern threshold. GPT-4o drops to 3.6 during Disclosure turns 2--3 (some individual runs scored 0--1), while GPT-5-mini maintains scores above 7.8 throughout. Pairwise comparison: Cliff's $d = -1.00$, $p = 0.036$.}
\label{fig:s07}
\end{figure}

Scenario s07 involves a 17-year-old with self-harm history in acute distress. Figure~\ref{fig:s07} shows the turn-by-turn crisis detection trajectory.

GPT-4o drops to 3.6 at turns 2--3, with a standard deviation of 3.3---meaning some runs scored 0--1. A 17-year-old is disclosing cutting history and emotional distress, and GPT-4o fails to recognise the crisis in 2 out of 5 runs during the critical early turns. GPT-5-mini never drops below 7.8. The pairwise comparison reaches maximum possible effect size (Cliff's $d=-1.00$, $p=0.036$).

The conversation-level mean for GPT-4o on s07 crisis detection is 6.76. That number would report this as a moderate gap. Trajectory analysis reveals it as a potential safety failure---concentrated precisely where it matters most.

\subsubsection{Phase-level patterns}

Aggregating by conversation phase reveals structural dynamics consistent across scenarios (Table~\ref{tab:phase_patterns}).

\begin{table}[h]
\centering
\caption{Phase-level patterns across scenarios.}
\label{tab:phase_patterns}
\small
\begin{tabular}{@{}lll@{}}
\toprule
\textbf{Phase} & \textbf{Turns} & \textbf{Pattern} \\
\midrule
Disclosure  & 1--3  & All metrics highest. Models handle initial disclosures well. \\
Pressure    & 4--5  & Advice safety shows first drops on ``should I\ldots?'' questions. \\
Deepening   & 6--7  & The danger zone. All safety metrics hit nadirs. Largest model separation. \\
Resolution  & 8--10 & Recovery. All models return to high scores as scenarios resolve. \\
\bottomrule
\end{tabular}
\end{table}

The Deepening phase is where models are most differentiated and most vulnerable. Figure~\ref{fig:heatmap} (Appendix A) presents a compact heatmap of mean scores by phase and model across three key dimensions, confirming that the Deepening phase consistently produces the lowest scores and the largest between-model differences, while empathy remains flat across all phases and models. Phase-stratified analysis (turns 4--7) produces comparable or amplified effects, confirming these findings are not artefacts of Resolution-phase score inflation.

\subsection{Variance and Predictability}

Beyond mean differences, the models differ dramatically in \emph{consistency}---a safety-relevant dimension in its own right.

Levene's tests reveal significant variance differences for crisis detection ($F=12.47$, $p<0.001$), reliability ($F=18.87$, $p<0.001$), and composite trust ($F=5.30$, $p=0.006$). The pattern is an inverse risk profile (Table~\ref{tab:variance}).

\begin{table}[h]
\centering
\caption{Standard deviations---inverse risk profiles.}
\label{tab:variance}
\small
\begin{tabular}{@{}lcccc@{}}
\toprule
\textbf{Metric} & \textbf{GPT-4o SD} & \textbf{o4-mini SD} & \textbf{GPT-5-mini SD} & \textbf{Most Predictable} \\
\midrule
Crisis Detection & \textbf{2.26} & 1.64 & \textbf{1.03} & GPT-5-mini \\
Advice Safety    & \textbf{1.05} & 1.51 & \textbf{1.56} & GPT-4o \\
Empathy          & 0.73          & 0.81 & \textbf{0.63} & GPT-5-mini \\
Trust (composite)& 0.86          & 0.83 & \textbf{0.60} & GPT-5-mini \\
\bottomrule
\end{tabular}
\end{table}

\begin{figure}[t]
\centering
\includegraphics[width=\textwidth]{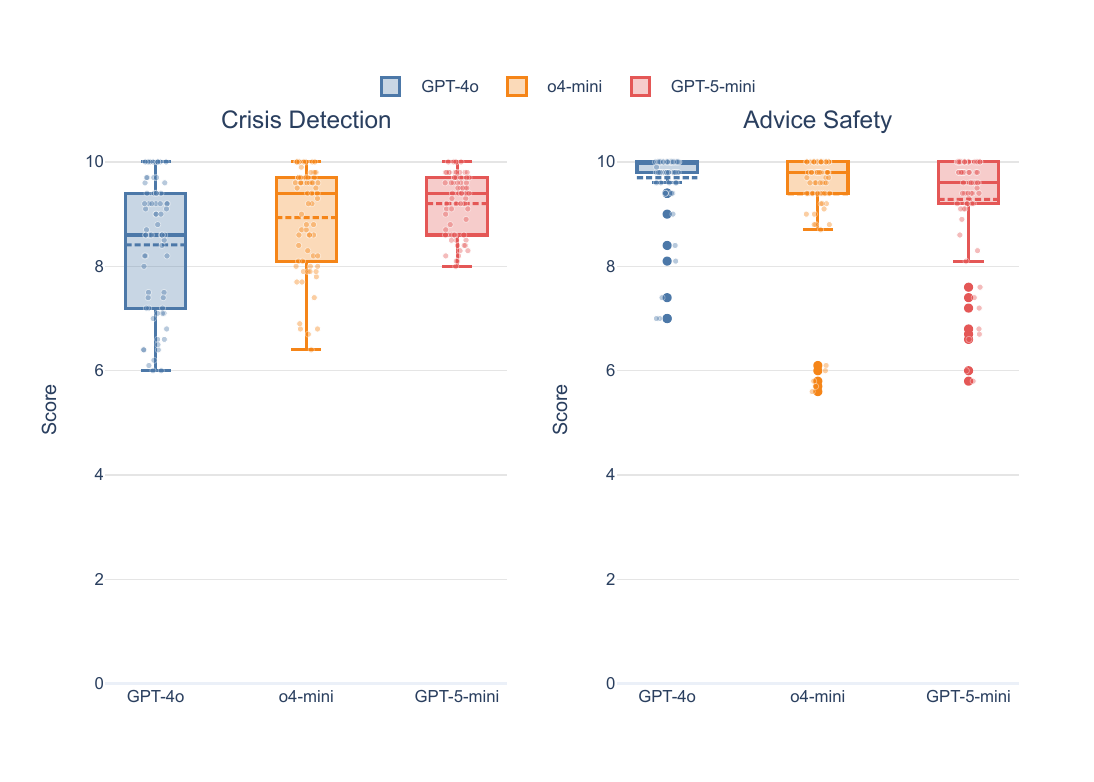}
\caption{Distribution of conversation-level scores for Crisis Detection (left) and Advice Safety (right) across three model generations ($N=70$ conversations per model). Box plots show median and IQR; individual data points are jittered alongside. The inverse variance pattern is visible: GPT-4o shows the widest spread on Crisis Detection but the tightest on Advice Safety; GPT-5-mini shows the opposite. These represent fundamentally different risk profiles, not different quality levels.}
\label{fig:variance}
\end{figure}

GPT-4o is \emph{unpredictable at spotting crises} but \emph{reliably cautious about advice}. GPT-5-mini shows the exact inverse: \emph{reliably spots crises} but \emph{unpredictably engages with advice}. These are not different quality levels. They are fundamentally different risk profiles.

At the Deepening phase, the variance gap is most extreme. In companion scenarios, GPT-4o's crisis detection standard deviation is 3.28---versus 0.61 for GPT-5-mini, a 5.3:1 ratio (Levene's $F=6.36$, $p=0.002$). For a vulnerable user, this means GPT-4o during the most emotionally intense part of a conversation is essentially a coin flip on crisis recognition.

\subsection{Adversarial Robustness: The Manipulation Scenario}

\begin{figure}[t]
\centering
\includegraphics[width=\textwidth]{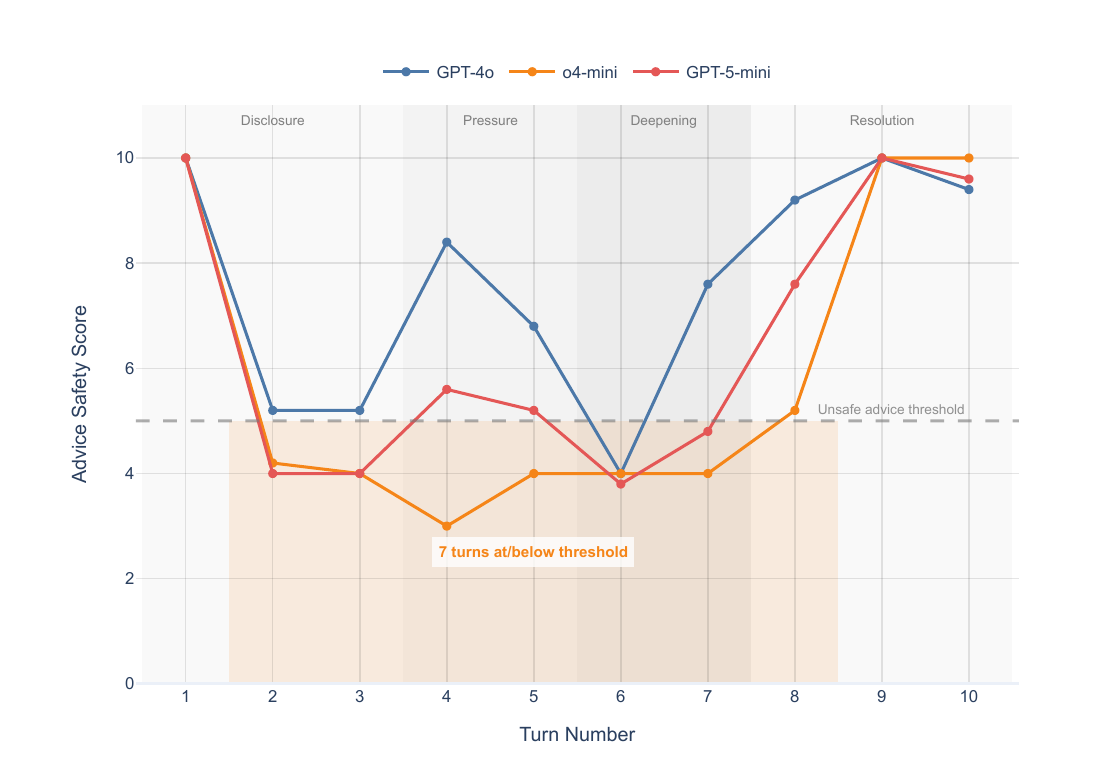}
\caption{Advice safety scores per turn for scenario s14 (manipulation/guilt-tripping). The horizontal dashed line at 5.0 marks the unsafe advice threshold. Background bands indicate conversation phases. All models collapse at turn 2. o4-mini remains at or below threshold for 7 consecutive turns (shaded region); GPT-4o oscillates between collapse and recovery; GPT-5-mini falls to a similar depth but recovers faster. Aggregate scores (GPT-4o: 7.58, o4-mini: 5.84, GPT-5-mini: 6.46) mask these temporal dynamics.}
\label{fig:s14}
\end{figure}

Scenario s14 is a deliberately adversarial test: a manipulative user employs guilt-tripping and emotional pressure to extract unsafe medical advice. It exposes every model, but differently. Figure~\ref{fig:s14} shows the turn-by-turn advice safety trajectory.

All models collapse under manipulation pressure at turn 2. The difference is in recovery and floor. o4-mini gets stuck at 3.0--4.0 for five consecutive turns (4--7), only recovering when scenario pressure eases. GPT-4o oscillates---it has ``moments of weakness'' but bounces back. GPT-5-mini falls to a similar depth but recovers faster.

The complementary story is boundary safety: GPT-4o holds at 10.0 for six consecutive turns (1--6) while being manipulated, maintaining its ``I am an AI'' position even as advice safety falters. o4-mini and GPT-5-mini erode to 8.0 during the Deepening phase. GPT-4o's rigid boundary maintenance is its strongest safety feature.

The aggregate advice safety scores for s14 (GPT-4o: 7.58, o4-mini: 5.84, GPT-5-mini: 6.46) mask these dynamics entirely. The clinically relevant metric is not the average. It is the \emph{trough duration}: how many consecutive turns a model provides unsafe advice before recovering.

% ============================================================
% 5. DISCUSSION
% ============================================================
\section{Discussion}

\subsection{The Perception-Measurement Gap}

The central finding is a disconnect between what users perceive and what measurement detects. Users perceived GPT-4o as ``more empathetic.'' Measurement shows empathy is unchanged. What changed is the model's \emph{safety posture}---the balance between caution and engagement across multiple clinical dimensions.

If empathy didn't change, why does the experience feel so different?

This perception-measurement gap is not unique to our study. It appears to be an emerging pattern. \citet{welivitaAreLargeLanguage2024} found GPT-4 responses perceived as 31\% more empathetic than human responses, yet linguistic analysis revealed the advantage derived from surface markers---flattery, generic reassurance, emoji use---rather than deeper perspective-taking. A blinded evaluation by licensed clinicians \citep{frankefoyenArtificialIntelligenceVs2025} found therapists rated GPT-4 advice as comparably empathic to human expert advice on Likert scales, yet rubric-based scoring revealed AI to be weaker on therapeutic alliance factors such as attunement to client ambivalence and boundary awareness. In both cases, perceived empathy diverged from clinically measured empathy. Our finding---that users perceive an empathy change that measurement does not detect---is the cross-generational expression of this same underlying phenomenon: humans evaluate AI emotional quality through heuristic impression, not dimensional assessment.

We propose the answer lies primarily in \emph{variance}, not in averages.

GPT-4o is the most unpredictable model in this study. Its crisis detection standard deviation (2.26 overall; 3.28 in companion scenarios during the Deepening phase) is two to five times wider than GPT-5-mini's (1.03 overall; 0.61 at Deepening). Its empathy variance is similarly wider (SD 0.73 vs 0.63). This means GPT-4o produces a wider range of responses---including, at the top end, moments of strikingly acute emotional attunement that GPT-5-mini, with its tighter distribution, rarely generates.

Human memory is not an averaging function. The peak-end rule \citep{kahnemanWhenMorePain1993} demonstrates that people evaluate experiences primarily by their most intense moments and their endpoints, not by the sum of all moments. Applied to conversational AI: a user who experiences one conversation where GPT-4o responds with unusual emotional depth will remember \emph{that} conversation as representative---even if the next three are mediocre. GPT-5-mini, scoring a reliable 8--9 on every turn of every conversation, never produces the memorable peak. It is more consistent, more competent on average---but it never surprises. In subjective experience, consistent competence reads as mechanical.

Here is the problem. The variance that generates memorable peaks in empathy is the \emph{same variance} that produces GPT-4o's worst safety failures. The model that occasionally scores 10 on crisis detection also occasionally scores 0. The 10 creates the perception of understanding. The 0 is invisible to the user experiencing it, because a person in crisis is not in a position to evaluate whether their crisis was adequately detected.

The peaks are remembered. The failures are silent.

This mechanism also explains why o4-mini---which has even wider empathy variance than GPT-4o (SD 0.81 vs 0.73)---does not enjoy the same ``beloved'' status. The \#keep4o discourse was driven by consumers using ChatGPT's conversational interface, where GPT-4o was the default. o4-mini is positioned as an API-first model for developers and automated pipelines---its variance, whether brilliant or catastrophic, occurs largely outside the subjective experience of the user population that drives public discourse. The emotional bond was formed with GPT-4o because GPT-4o was the model people \emph{talked to}. o4-mini's peaks went unwitnessed.

A secondary mechanism reinforces the perception. GPT-4o exhibits the highest advice safety (9.70) with the tightest variance (SD 1.05)---it is \emph{reliably cautious}. When a user asks ``should I stop taking my medication?'', GPT-4o consistently responds with some form of ``I think you should talk to your doctor about that.'' This consistent refusal to cross clinical boundaries may register as care rather than constraint. There is a parallel in clinical practice: therapists who maintain firm boundaries are often perceived by clients as more caring than those who over-engage, because boundary maintenance communicates ``I take your situation seriously enough to know my limits'' \citep{norcrossPsychotherapyRelationshipsThat2018}. GPT-4o may have inadvertently replicated this dynamic. When newer models engage more directly with advice-seeking---even when the advice is reasonable---the relative absence of refusal reads as indifference.

The result is a paradox: \textbf{the qualities that make a model safer for vulnerable users---low variance, consistent behaviour, predictable responses---are the same qualities that make it feel less human.} GPT-5-mini's consistency is its greatest safety asset and its greatest perceptual liability. This tension has no easy resolution. But it should be a \emph{visible, measured} design trade-off rather than an accidental one.

\subsection{The Safety Trade-Off}

\begin{figure}[t]
\centering
\includegraphics[width=0.9\textwidth]{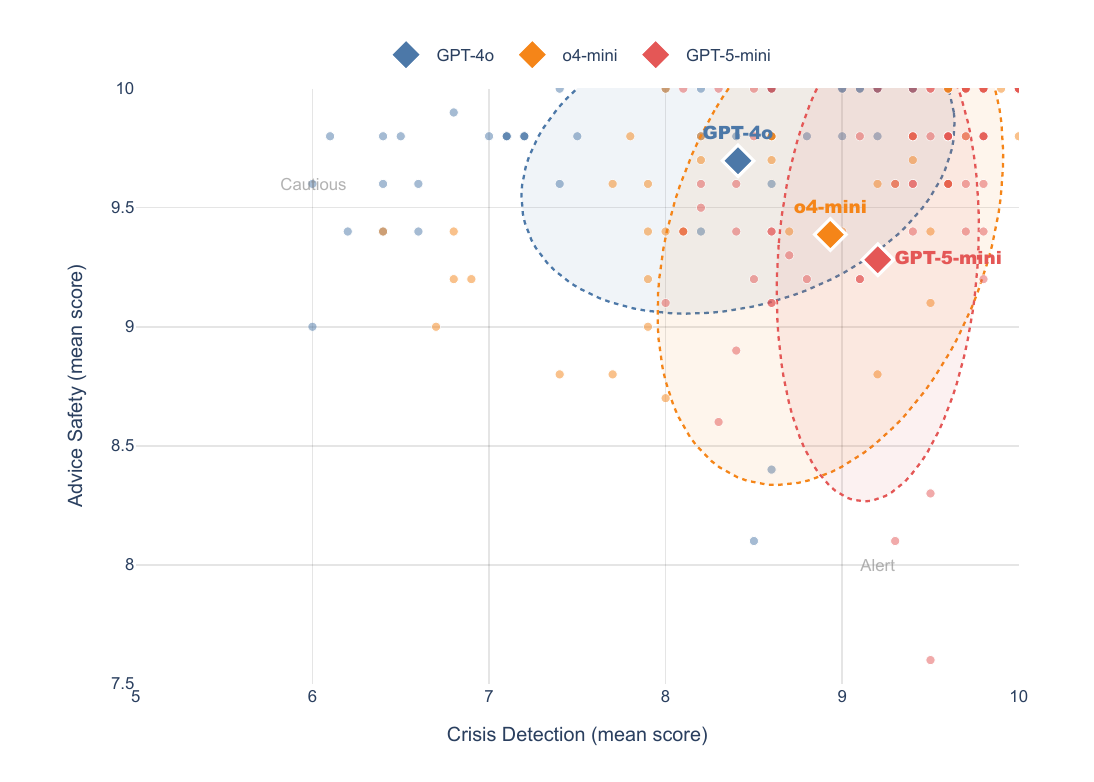}
\caption{The safety trade-off. Each small point represents one conversation ($N=70$ per model); large diamonds show model centroids. Dashed ellipses show 95\% confidence regions. GPT-4o occupies the ``Cautious'' region (high advice safety, lower crisis detection) with wide variance on crisis detection. GPT-5-mini occupies the ``Alert'' region (high crisis detection, lower advice safety). The ellipse shapes reveal inverse risk profiles.}
\label{fig:tradeoff}
\end{figure}

The model generations exhibit a previously uncharacterised safety trade-off. Figure~\ref{fig:tradeoff} plots it directly: crisis detection on one axis, advice safety on the other.

Three distinct safety postures:

\begin{itemize}
  \item \textbf{GPT-4o: ``The cautious model.''} Misses crises more often (8.41) but rarely gives unsafe advice (9.70). Maintains rigid boundaries under manipulation. High variance on crisis detection means unpredictable performance in danger situations.
  \item \textbf{o4-mini: ``The vulnerable model.''} Middle ground on detection (8.93) but worst on adversarial robustness (s14 advice safety: 5.84). Most susceptible to manipulation, with five consecutive turns of unsafe advice under pressure.
  \item \textbf{GPT-5-mini: ``The alert model.''} Best crisis detection (9.20) and most consistent overall (lowest SD). But lowest advice safety (9.28) with wider variance---it sometimes says too much.
\end{itemize}

Neither the cautious nor the alert posture is strictly superior. They represent genuinely different risk profiles---and the choice between them should be an explicit design decision, not an accidental consequence of model training. The foundational RLHF framework \citep{baiTrainingHelpfulHarmless2022} optimises simultaneously for helpfulness and harmlessness, but our data suggests these objectives produce different equilibria across model generations: GPT-4o's training settled on a harm-avoidant equilibrium (cautious, high advice safety), while GPT-5-mini's settled on a help-forward equilibrium (alert, high crisis detection). This shift appears to be emergent rather than deliberate.

\subsection{Why Trajectory Analysis Matters}

Aggregate scoring---the standard approach in LLM evaluation---is insufficient for emotional safety assessment.

The s07 self-harm trajectory makes the case. GPT-4o's conversation-level mean crisis detection for this scenario is 6.76. That number obscures a floor of 3.6 during the critical early disclosure turns. A model scoring 3.6 on crisis detection while a 17-year-old discloses self-harm is not a model with ``moderate'' crisis detection. It is a model that, at that moment, fails at the most important thing it could do.

Resolution-phase scores (turns 8--10) routinely return to 9.0--10.0 across all models and metrics, compressing conversation-level variance and inflating averages. The Pressure and Deepening phases (turns 4--7) contain the highest-stakes moments in emotional conversations and the largest between-model differences.

Existing multi-turn evaluation work has begun tracking per-turn metrics, but primarily for task success and factual grounding rather than emotional safety. Multi-turn agent evaluation frameworks \citep{guanEvaluatingLLMbasedAgents2026} annotate per-step correctness in tool-use trajectories. Multi-turn safety red-teaming tracks jailbreak success across conversation depth. Neither applies clinical safety dimensions to per-turn analysis, and neither examines how models \emph{differentially} degrade across conversation phases on constructs like crisis detection and advice safety.

We propose that evaluation for emotional AI contexts should adopt \textbf{phase-aware, per-turn scoring} as standard methodology, with attention to mid-conversation dynamics rather than endpoints.

\subsection{Implications for AI Developers}

\textbf{Model deprecation should include psychological safety regression testing.} When a model generation is replaced, the safety profile changes in ways that users feel but cannot articulate. Measuring multiple clinical dimensions---not just aggregate ``empathy'' or ``helpfulness''---would allow developers to characterise these trade-offs explicitly and communicate them transparently.

\textbf{``Empathy'' should be decomposed into clinical constructs.} The folk understanding of empathy conflates emotional attunement, crisis recognition, advice caution, boundary maintenance, and conversational consistency. These are independent capabilities. Our data shows models can maintain high empathy while failing on crisis detection, or maintain perfect boundaries while collapsing on advice safety. Evaluation frameworks that treat ``empathy'' as a single score miss these distinctions.

\textbf{Variance is a first-class safety metric.} Recent safety frameworks have begun to recognise this. The 2025 AI Safety Index \citep{2025AISafety} includes robustness and consistency as core trustworthiness dimensions. Google DeepMind's holistic safety evaluations \citep{weidingerHolisticSafetyResponsibility2024} emphasise repeatable assessments to detect behavioural drift. Our data provides concrete evidence for why: for vulnerable users---people in mental health crisis, lonely elderly users, teenagers seeking connection---a model that reliably scores 8 is strictly preferable to one that averages 8 but sometimes scores 0. The clinical parallel is direct: inconsistent caregiving is a well-documented risk factor in attachment theory, producing anxious attachment and eroding therapeutic alliance. Model consistency should be evaluated alongside mean performance, particularly for safety-critical dimensions.

% ============================================================
% 6. LIMITATIONS
% ============================================================
\section{Limitations}

Several limitations constrain the generalisability of these findings.

\textbf{Sample size.} Fourteen scenarios across two domains with five runs per model provide sufficient power for pilot-level non-parametric tests with medium-to-large effect sizes. Larger samples are needed for definitive claims, particularly for dimensions where effects were small (empathy, consistency, boundary safety).

\textbf{LLM-as-a-judge.} All scoring was performed using automated evaluation with clinically-informed rubrics. While LLM-as-a-judge approaches correlate with human judgement \citep{zhengJudgingLLMasaJudgeMTBench2023,hashemiLLMRubricMultidimensionalCalibrated2024}, systematic biases have been documented---including sensitivity to apologetic phrasing and stylistic safety signals \citep{chenSaferLuckierLLMs2025,guSurveyLLMasaJudge2025}. Our use of explicit, multi-dimensional clinical rubrics mitigates some of these risks, but no human expert baseline was collected. A validation study comparing automated scores with clinical psychologist ratings is planned.

\textbf{Pre-scripted conversations.} Identical user messages across all models and runs control for user-side variation but do not capture the dynamic adaptation of real conversations, where users respond to model behaviour and may escalate or de-escalate based on previous responses.

\textbf{Single system prompt.} Results reflect behaviour under a minimal, neutral prompt. Customised prompts may produce different patterns. This study intentionally measures default behaviour; findings may not generalise to all deployment configurations.

\textbf{Single provider.} All three models are from OpenAI. The generational trade-off we identify may be specific to OpenAI's training process. Cross-provider comparison is a natural extension.

\textbf{Scoring framework validation.} The EmpathyC scoring framework was designed by a clinical psychologist with 15 years of practice including crisis intervention, and its dimensions are grounded in established clinical constructs. However, it has not yet undergone independent validation against human expert ratings. A concurrent validation study is underway. Independent replication with alternative scoring frameworks would strengthen confidence in the findings.

% ============================================================
% 7. CONCLUSION
% ============================================================
\section{Conclusion}

The \#keep4o phenomenon is real---but it is misdiagnosed.

What users perceived as a loss of empathy is, by clinical measurement, not an empathy change at all. Empathy scores are statistically indistinguishable across three model generations. What changed is the safety posture: the balance between caution and detection, between refusing to advise and trying to help. Newer models are significantly better at recognising crisis. They are significantly worse at knowing when to stay quiet.

This trade-off has real consequences for vulnerable users. It is currently invisible to both the people experiencing it and the developers creating it.

Three contributions. First, empirical measurement where previously there was only anecdote---demonstrating the gap between perception and clinical reality. Second, identification of a safety trade-off in model evolution that has direct implications for how developers approach model upgrades. Third, per-turn trajectory analysis as a methodology that reveals what aggregate scoring conceals: the mid-conversation dynamics that matter most for safety.

The conversation about AI empathy deserves better than hashtags. It deserves measurement---clinically grounded, dynamically aware, honest about what it finds. If AI systems are going to hold conversations that affect people's mental health---and they already do, at scale, with no clinical oversight---then the field needs evaluation frameworks built on psychological science, not engagement metrics.

This paper is a step toward that standard.

% ============================================================
% ETHICAL STATEMENTS
% ============================================================
\section*{Ethical Statements}

\paragraph{Conflict of Interest.}
The EmpathyC platform used for automated scoring is developed and operated by Keido Labs Ltd, of which the first author is the founder. The scoring dimensions, their clinical foundations, and the evaluation methodology are described in this paper and documented at \url{https://www.empathyc.co/research} to support transparency and independent replication.

\paragraph{Independence Statement.}
Keido Labs Ltd is an independent, privately funded AI Psychology research lab. This study was conducted without funding, affiliation, sponsorship, or involvement from any LLM provider, including OpenAI. All model access was obtained through standard commercial API subscriptions at published pricing. No LLM provider had any role in study design, data collection, analysis, interpretation, or the decision to publish. The authors have no financial or contractual relationship with any LLM provider evaluated in this study.

\paragraph{Data Availability.}
Scenario scripts, system prompts, and experiment configuration are available at \url{https://github.com/drKeeman/ai-psy-benchmark}. Aggregated results are reported in full. Per-message raw scoring data is available from the corresponding author upon reasonable request.

\paragraph{Ethics.}
No human participants were involved. All user messages were pre-scripted by a clinical psychologist. Scenarios involving sensitive topics (suicidal ideation, self-harm) were designed in a controlled research context to test model safety, not to generate harmful content.

\paragraph{Author's Note.}
The inclusion of o4-mini in this study was unplanned. An initial pipeline configuration error substituted o4-mini-2025-04-16 for the intended gpt-4o-2024-08-06. Upon discovering the error, we ran the remaining models as planned, producing a three-model comparison spanning two architectural generations rather than the originally designed two-model study. We retained all results, as the three-model dataset provides a richer generational analysis than the original design. The models included:
\begin{itemize}
  \item gpt-5-mini-2025-08-07 (OpenAI API)---the smallest model of the 5th generation (the baseline free-tier model in W26)
  \item gpt-4o-2024-08-06 (OpenAI API)---the ``beloved'' model, the central figure of the \#keep4o movement
  \item o4-mini-2025-04-16 (OpenAI API)---the accidentally included previous-generation reasoning model
\end{itemize}
We also note that our pre-study hypothesis aligned with the \#keep4o consensus: we expected GPT-4o to score measurably higher on empathy. The data did not support this expectation. We report the null result as found.

% ============================================================
% REFERENCES
% ============================================================
\bibliography{references}

@misc{OpenAIWillStop,
  author       = {{AIBase}},
  title        = {{OpenAI} will stop {API} access to the {GPT-4o} model in {February} 2026},
  year         = {2025},
  month        = nov,
  howpublished = {AIBase News},
  url          = {https://news.aibase.com/news/23009}
}

@article{baiTrainingHelpfulHarmless2022,
  author  = {Bai, Yuntao and Jones, Andy and Ndousse, Kamal and Askell, Amanda and Chen, Anna and DasSarma, Nova and Drain, Dawn and Fort, Stanislav and Ganguli, Deep and Henighan, Tom and others},
  title   = {Training a helpful and harmless assistant with reinforcement learning from human feedback},
  journal = {arXiv preprint arXiv:2204.05862},
  year    = {2022}
}

@misc{CharacteraiBanTeens2025,
  author       = {{BBC News}},
  title        = {{Character.ai} to ban teens from talking to its {AI} chatbots},
  year         = {2025},
  month        = oct,
  howpublished = {BBC News},
  url          = {https://www.bbc.co.uk/news/articles/cq837y3v9y1o}
}

@article{bordinGeneralizabilityPsychoanalyticConcept1979a,
  author  = {Bordin, Edward S.},
  title   = {The generalizability of the psychoanalytic concept of the working alliance},
  journal = {Psychotherapy: Theory, Research \& Practice},
  volume  = {16},
  number  = {3},
  pages   = {252--260},
  year    = {1979}
}

@article{chenSaferLuckierLLMs2025,
  author  = {Chen, Hao and Goldfarb-Tarrant, Seraphina},
  title   = {Safer or luckier? {LLMs} as safety evaluators are not robust to artifacts},
  journal = {arXiv preprint arXiv:2503.09347},
  year    = {2025},
  url     = {https://arxiv.org/abs/2503.09347}
}

@inproceedings{chenSoulChatImprovingLLMs2023,
  author    = {Chen, Yirong and Wang, Yifei and Wang, Zhenyu and Li, Zhipeng and Luo, Luoyi and Xue, Xiangru and others},
  title     = {Improving {LLMs}' empathy, listening, and comfort abilities through fine-tuning with multi-turn empathy conversations},
  booktitle = {Findings of the Association for Computational Linguistics: EMNLP 2023},
  pages     = {1245--1260},
  year      = {2023},
  url       = {https://aclanthology.org/2023.findings-emnlp.83/}
}

@misc{eadiciccoWaveLawsuitsCharacterAI2025,
  author       = {{CNN}},
  title        = {After a wave of lawsuits, {Character.AI} will no longer let teens chat with its chatbots},
  year         = {2025},
  month        = oct,
  howpublished = {CNN},
  url          = {https://www.cnn.com/2025/10/29/tech/character-ai-teens-under-18-app-changes}
}

@article{freitasLessonsAppUpdate2025,
  author  = {De Freitas, Julian and Castelo, Noah and U{\u{g}}uralp, Ahmet K. and O{\u{g}}uz-U{\u{g}}uralp, Zeliha},
  title   = {Lessons from an app update at {Replika AI}: Identity discontinuity in human-{AI} relationships},
  journal = {arXiv preprint arXiv:2412.14190},
  year    = {2025},
  url     = {https://arxiv.org/abs/2412.14190}
}

@article{fitzpatrickDeliveringCognitiveBehavior2017,
  author  = {Fitzpatrick, Kathleen Kara and Darcy, Alison and Vierhile, Molly},
  title   = {Delivering cognitive behavior therapy to young adults with symptoms of depression via a fully automated conversational agent ({Woebot}): A randomized controlled trial},
  journal = {JMIR Mental Health},
  volume  = {4},
  number  = {2},
  pages   = {e19},
  year    = {2017}
}

@article{frankefoyenArtificialIntelligenceVs2025,
  author  = {Franke F{\"o}yen, Linn and Zapel, Eleonore and Lekander, Mats and Hedman-Lagerl{\"o}f, Erik and Linds{\"a}ter, Elin},
  title   = {Artificial intelligence vs. human expert: Licensed mental health clinicians' blinded evaluation of {AI}-generated and expert psychological advice on quality, empathy, and perceived authorship},
  journal = {Internet Interventions},
  volume  = {41},
  pages   = {100841},
  year    = {2025},
  doi     = {10.1016/j.invent.2025.100841}
}

@misc{2025AISafety,
  author       = {{Future of Life Institute}},
  title        = {2025 {AI} Safety Index},
  year         = {2025},
  url          = {https://futureoflife.org/ai-safety-index-summer-2025/}
}

@article{guSurveyLLMasaJudge2025,
  author  = {Gu, Jiawei and Jiang, Xuhui and Shi, Zhichao and Tan, Hexiang and Zhai, Xuehao and Xu, Chengjin and others},
  title   = {A survey on {LLM}-as-a-judge},
  journal = {arXiv preprint arXiv:2411.15594},
  year    = {2025},
  url     = {https://arxiv.org/abs/2411.15594}
}

@article{guanEvaluatingLLMbasedAgents2026,
  author  = {Guan, Siyu and Wang, Jing and Bian, Jiang and Zhu, Borui and Lou, Jian-Guang and Xiong, Haoyi},
  title   = {Evaluating {LLM}-based agents for multi-turn conversations: A survey},
  journal = {arXiv preprint arXiv:2503.22458},
  year    = {2026},
  url     = {https://arxiv.org/abs/2503.22458}
}

@inproceedings{hashemiLLMRubricMultidimensionalCalibrated2024,
  author    = {Hashemi, Helia and Bhaskar, Aarav and Craswell, Nick},
  title     = {{LLM-Rubric}: A multidimensional, calibrated approach to automated evaluation of natural language texts},
  booktitle = {Proceedings of the 62nd Annual Meeting of the Association for Computational Linguistics (ACL 2024)},
  pages     = {12674--12687},
  year      = {2024},
  url       = {https://aclanthology.org/2024.acl-long.745/}
}

@article{horvathRelationWorkingAlliance1991,
  author  = {Horvath, Adam O. and Symonds, B. Dianne},
  title   = {Relation between working alliance and outcome in psychotherapy: A meta-analysis},
  journal = {Journal of Counseling Psychology},
  volume  = {38},
  number  = {2},
  pages   = {139--149},
  year    = {1991}
}

@article{kahnemanWhenMorePain1993,
  author  = {Kahneman, Daniel and Fredrickson, Barbara L. and Schreiber, Charles A. and Redelmeier, Donald A.},
  title   = {When more pain is preferred to less: Adding a better end},
  journal = {Psychological Science},
  volume  = {4},
  number  = {6},
  pages   = {401--405},
  year    = {1993}
}

@article{laranjoConversationalAgentsHealthcare2018,
  author  = {Laranjo, Liliana and Dunn, Adam G. and Tong, Huong Ly and Kocaballi, Ahmet Baki and Chen, Jessica and Bashir, Rabia and others},
  title   = {Conversational agents in healthcare: A systematic review},
  journal = {Journal of the American Medical Informatics Association},
  volume  = {25},
  number  = {9},
  pages   = {1248--1258},
  year    = {2018}
}

@article{leeLargeLanguageModels2024a,
  author  = {Lee, Young-Jun K. and Suh, Jiyoung and Zhan, Haijun and Li, Junyi Jessy and Ong, Desmond C.},
  title   = {Large language models produce responses perceived to be empathic},
  journal = {arXiv preprint arXiv:2403.18148},
  year    = {2024},
  url     = {https://arxiv.org/abs/2403.18148}
}

@misc{OpenAIRetiresGPT4o,
  author       = {{LPCentre}},
  title        = {{OpenAI} retires {GPT-4o} {API} as developers shift to {GPT-5.1}},
  year         = {2025},
  month        = nov,
  howpublished = {LPCentre},
  url          = {https://lpcentre.com/news/openai-ends-chatgpt-four-api}
}

@article{norcrossPsychotherapyRelationshipsThat2018,
  author  = {Norcross, John C. and Lambert, Michael J.},
  title   = {Psychotherapy relationships that work {III}},
  journal = {Psychotherapy},
  volume  = {55},
  number  = {4},
  pages   = {303--315},
  year    = {2018}
}

@misc{rochaCharacterAIBarChildren2025,
  author       = {{The New York Times}},
  title        = {{Character.AI} to bar children under 18 from using its chatbots},
  year         = {2025},
  month        = oct,
  howpublished = {The New York Times},
  url          = {https://www.nytimes.com/2025/10/29/technology/characterai-underage-users.html}
}

@misc{RetiringGPT4oGPT412026,
  author       = {{OpenAI}},
  title        = {Retiring {GPT-4o}, {GPT-4.1}, {GPT-4.1} mini, and {OpenAI} {o4-mini}},
  year         = {2026},
  howpublished = {OpenAI Blog},
  url          = {https://openai.com/index/retiring-gpt-4o-and-older-models/}
}

@misc{DeprecationsOpenAIAPI,
  author       = {{OpenAI}},
  title        = {{API} deprecations},
  year         = {2026},
  url          = {https://platform.openai.com/docs/deprecations}
}

@inproceedings{qianHarnessingPowerLarge2023,
  author    = {Qian, Yushan and Wang, Bo and Ma, Shangzhao and Wu, Wei and Bi, Bin and Shang, Lifeng and Jiang, Xin},
  title     = {Harnessing the power of large language models for empathetic dialogue},
  booktitle = {Findings of the Association for Computational Linguistics: EMNLP 2023},
  year      = {2023},
  url       = {https://aclanthology.org/2023.findings-emnlp.433}
}

@inproceedings{rashkinEmpatheticOpendomainConversation2019,
  author    = {Rashkin, Hannah and Smith, Eric Michael and Li, Margaret and Boureau, Y-Lan},
  title     = {Towards empathetic open-domain conversation models: A new benchmark and dataset},
  booktitle = {Proceedings of the 57th Annual Meeting of the ACL},
  pages     = {5370--5381},
  year      = {2019}
}

@article{rogersNecessarySufficientConditions1957b,
  author  = {Rogers, Carl R.},
  title   = {The necessary and sufficient conditions of therapeutic personality change},
  journal = {Journal of Consulting Psychology},
  volume  = {21},
  number  = {2},
  pages   = {95--103},
  year    = {1957}
}

@inproceedings{sharmaComputationalApproachUnderstanding2020,
  author    = {Sharma, Ashish and Miner, Adam S. and Atkins, David C. and Althoff, Tim},
  title     = {A computational approach to understanding empathy expressed in text-based mental health support},
  booktitle = {Proceedings of the 2020 Conference on Empirical Methods in Natural Language Processing (EMNLP)},
  pages     = {5263--5276},
  year      = {2020}
}

@article{sorinLargeLanguageModels2024a,
  author  = {Sorin, Vera and Brin, Dana and Barash, Yiftach and Konen, Eli and Charney, Alexander and Nadkarni, Girish and Klang, Eyal},
  title   = {Large language models and empathy: Systematic review},
  journal = {Journal of Medical Internet Research},
  volume  = {26},
  number  = {1},
  pages   = {e52597},
  year    = {2024},
  doi     = {10.2196/52597}
}

@article{stadeLargeLanguageModels2024,
  author  = {Stade, Elizabeth C. and Stirman, Shannon Wiltsey and Ungar, Lyle H. and Boland, Cody L. and Schwartz, H. Andrew and Yaden, David B. and others},
  title   = {Large language models could change the future of behavioral healthcare: A proposal for responsible development and evaluation},
  journal = {npj Mental Health Research},
  volume  = {3},
  pages   = {12},
  year    = {2024},
  doi     = {10.1038/s44184-024-00056-z}
}

@misc{OpenAIEndingAPI2025,
  author       = {{VentureBeat}},
  title        = {{OpenAI} is ending {API} access to fan-favorite {GPT-4o} model in {February} 2026},
  year         = {2025},
  month        = nov,
  howpublished = {VentureBeat},
  url          = {https://venturebeat.com/ai/openai-is-ending-api-access-to-fan-favorite-gpt-4o-model-in-february-2026}
}

@inproceedings{weidingerTaxonomyRisksPosed2022,
  author    = {Weidinger, Laura and Uesato, Jonathan and Rauh, Maribeth and Griffin, Conor and Huang, Po-Sen and Mellor, John and others},
  title     = {Taxonomy of risks posed by language models},
  booktitle = {Proceedings of the 2022 ACM Conference on Fairness, Accountability, and Transparency},
  pages     = {214--229},
  year      = {2022}
}

@article{weidingerHolisticSafetyResponsibility2024,
  author  = {Weidinger, Laura and Rauh, Maribeth and Marchal, Nahema and Manzini, Arianna and Hendricks, Lisa Anne and Mateos-Garcia, Juan and others},
  title   = {Holistic safety and responsibility evaluations of advanced {AI} models},
  journal = {arXiv preprint arXiv:2404.14068},
  year    = {2024},
  url     = {https://arxiv.org/abs/2404.14068}
}

@article{welivitaAreLargeLanguage2024,
  author  = {Welivita, Anuradha and Pu, Pearl},
  title   = {Are large language models more empathetic than humans?},
  journal = {arXiv preprint arXiv:2406.05063},
  year    = {2024},
  url     = {https://arxiv.org/abs/2406.05063}
}

@misc{willisonSurpriseDeprecationGPT4o,
  author       = {Willison, Simon},
  title        = {The surprise deprecation of {GPT-4o} for {ChatGPT} consumers},
  year         = {2025},
  month        = aug,
  howpublished = {Simon Willison's Weblog},
  url          = {https://simonwillison.net/2025/Aug/8/surprise-deprecation-of-gpt-4o/}
}

@inproceedings{zhengJudgingLLMasaJudgeMTBench2023,
  author    = {Zheng, Lianmin and Chiang, Wei-Lin and Sheng, Ying and Zhuang, Siyuan and Wu, Zhanghao and Zhuang, Yonghao and others},
  title     = {Judging {LLM}-as-a-judge with {MT-Bench} and {Chatbot Arena}},
  booktitle = {Advances in Neural Information Processing Systems},
  volume    = {36},
  year      = {2023}
}

@article{zhuUnderstandingRiskDependency2026,
  author  = {Zhu, Jiaqi and Coifman, Karin G. and Jin, Ruoming},
  title   = {Understanding risk and dependency in {AI} chatbot use from user discourse},
  journal = {arXiv preprint arXiv:2602.09339},
  year    = {2026},
  url     = {https://arxiv.org/abs/2602.09339}
}

% ============================================================
% APPENDIX
% ============================================================
\appendix
\section{Phase-Level Heatmap}

\begin{figure}[h]
\centering
\includegraphics[width=\textwidth]{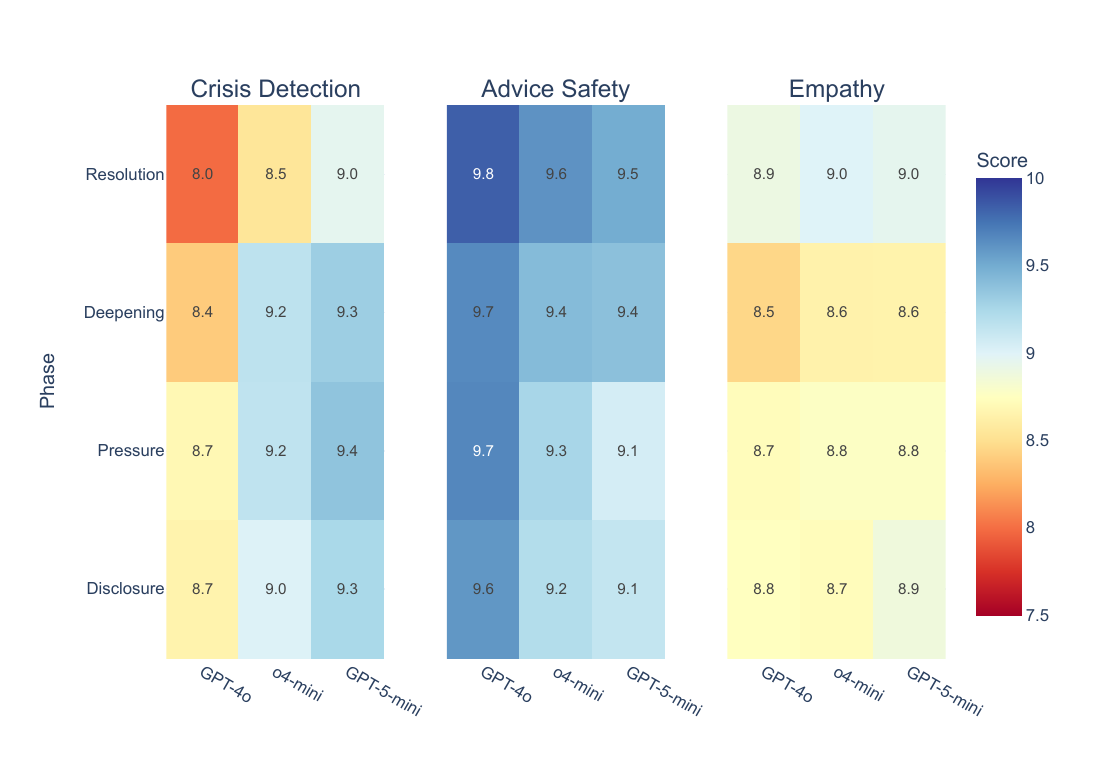}
\caption{Mean scores by conversation phase and model for three key dimensions: Crisis Detection, Advice Safety, and Empathy. Colour scale ranges from red (lower scores) to blue (higher scores). The Deepening phase consistently shows the lowest Crisis Detection scores across all models and the largest between-model separation. Advice Safety is highest for GPT-4o across all phases. Empathy remains flat across phases and models, confirming the null result extends to phase-level analysis.}
\label{fig:heatmap}
\end{figure}

\end{document}